# Multiple Leak Localization in Water Distribution Networks through Graph Neural Networks

**Arvin Ajoodani[1], Sara Nazif[1], and Pouria Ramazi[2]**

1. School of Civil Engineering, College of Engineering, University of Tehran, Tehran, Iran
2. Department of Biological Sciences and Department of Mathematics and Statistics, University of Calgary, Calgary, Canada

## Abstract

Leak localization in water distribution networks is challenging when multiple leaks occur simultaneously, particularly under sparse monitoring. Existing data-driven methods often assume a single or predefined number of leaks or require extensive enumeration of leak combinations. This study proposes a graph neural network (GNN) model that localizes a variable number of simultaneous leaks without specifying the leak count in advance or exhaustively generating every possible combination of leak locations for training. The model is evaluated on two benchmark networks of different sizes, the 31-node Hanoi network and the larger 92-node NET3 network, under variable demand, sparse pressure-sensor coverage, variable leak magnitudes, and varying numbers of simultaneous leaks. The GNN outperforms multilayer perceptron, support vector machine, K-nearest neighbors, and random forest classifiers by up to 12% and 20% in accuracy respectively in Hanoi and NET3 networks. Consistent performance across both networks supports the scalability of the proposed approach.

## 1. Introduction

Leakage in water distribution networks (WDNs) is a major global challenge. Global physical water losses in WDNs have been estimated at approximately 32 billion m³ per year, with about half occurring in developing countries (AbuEltayef et al., 2023). These losses threaten water security and impose substantial financial and environmental costs on communities (Fang et al., 2019). Reducing avoidable distribution losses supports Sustainable Development Goal 6 by improving water-use efficiency and the reliability of urban water services (Callejas Moncaleano et al., 2021). Conventional leakage-management and model-based localization methods are sensitive to monitoring availability and hydraulic-model accuracy (Puust et al., 2010; Marzola et al., 2022).

Data-driven approaches have emerged as a promising alternative, using machine learning (ML) techniques to detect and localize leaks from hydraulic data. Recent studies have demonstrated the potential of Gaussian naive Bayes classifiers (Mazaev et al., 2022), neural networks (Sun et al., 2020), and random forests (RF) (Lučin et al., 2021) for leak localization in WDNs. However, most of these studies assumed the presence of a single leak, which does not reflect conditions in many WDNs. Aging infrastructure and continued urbanization increase pipe deterioration and leakage risk, while climatic factors such as temperature, moisture, and precipitation can contribute to water-main deterioration; together, these pressures can increase the likelihood of concurrent leaks in large WDNs (Ahmad et al., 2023; Chen et al., 2026).

A few studies have addressed multiple leak localization using data-driven approaches. For example, Rodríguez-Argote et al. (2023) employed classifiers to isolate two simultaneous leaking zones in a WDN, while Fang et al. (2019) used convolutional neural networks to detect multiple leaks in an experimental network. Although these studies represent important steps forward, they are designed for a fixed number of leaks or a limited set of leak-location configurations. Exhaustively generating training scenarios for every possible combination of multiple leaks becomes computationally prohibitive as the number of candidate nodes increases, and models designed for a fixed leak count may not generalize to variable leakage conditions (Romero-Ben et al., 2023).

Graph-based methods provide a way to address the two limitations identified above: handling different numbers of simultaneous leaks and using limited hydraulic measurements without testing every possible combination of leak location. WDNs are naturally suited to graph-based modeling, with nodes representing consumption nodes and edges representing the pipes that connect them. Graph neural networks (GNN) can capture spatial dependencies in graph-structured data (Xia et al., 2021) and have been applied in fields such as finance (Wang et al., 2022), biology (Zhang et al., 2021), and electrical engineering (Liao et al., 2022). This makes GNN a promising approach for leak detection and localization in WDNs (Yang et al., 2022). By incorporating network topology into message passing, GNN can learn topology-dependent hydraulic relationships relevant to leak localization (Garðarsson et al., 2022).

Several recent studies have used GNN for leak localization in WDNs. Şahin and Yüce (2023) used graph convolutional networks (GCN) to classify pipeline conditions as leak, risky, or normal and reported better results than a support vector machine (SVM). Rolle et al. (2024) evaluated GNN-based localization when only limited sensor data were available and showed that useful location information could still be obtained after reconstructing missing measurements. Li and Wu (2024) combined node information with pipe information and localized leaks at the pipe level. Their study was based on single-leak cases and showed that the model could still provide useful localization when the number of pressure sensors was reduced. Zhang and Fink (2026) proposed an algorithm-informed GNN that estimates the current pressure field and the pressure field expected without leakage. Leaks are then detected and localized from the difference between these two estimates.

More recent studies have started to consider simultaneous leaks under different choices of leak counts and candidate locations.. Li et al. (2025) evaluated single, double, triple, and four-leak cases. In their small simulated network, four possible leaking pipes were selected first, and all combinations of one to four leaks among these pipes were generated. Separate databases were then created for the different numbers of leaks. In the L-Town test, the abnormal class also included cases with one or two leaks occurring at the same time. Yin et al. (2026) focused specifically on two simultaneous leaks and designed the model to return two leaking pipes and their positions. Kim and Choi (2026) considered one to four simultaneous leaks using a graph attention network. They avoided generating every possible combination by sampling the training data, but the maximum number of leaks was still fixed at four.

Taken together, the previous GNN studies have three main limitations. First, many studies consider only one leak at a time. Second, studies that consider several leaks usually define the number of leaks in advance, for example exactly two leaks or a range from one to four. Third, some studies build the data around a limited set of possible leak locations and explicitly generate the combinations among them. This becomes difficult for a large WDN because the number of possible combinations increases very quickly as the number of candidate nodes increases.

The proposed model performs node-level multilabel classification, allowing the number of predicted leaks to vary among scenarios. Consequently, the number of leaks need not be specified in advance, and the training set need not exhaustively enumerate all possible combinations of leaking nodes. The framework is evaluated under variable demand, sparse sensor coverage, and variable leak magnitudes on the Hanoi and NET3 networks. Furthermore, K-nearest neighbors (KNN), SVM, multilayer perceptron (MLP) and RF models were additionally developed to compare the proposed method with a broader range of ML classifiers.

## 2. Methodology

The proposed methodology consists of five main stages: leakage-scenario generation, construction of model inputs and target labels, development of the GCN and conventional classifiers, model training and evaluation, and application to benchmark WDNs.

### 2.1 Data Generation

The data-generation procedure is summarized in Figure 1. For each leakage scenario, the number of simultaneous leaks was first selected randomly from 1 to N, where N is the number of leak-candidate consumption nodes; reservoirs and tanks are treated as hydraulic boundary components and are not considered leak candidates. The number and locations of simultaneous leaks were sampled over a broad range to expose the models to diverse multilabel configurations and reduce potential class imbalance, rather than to reproduce the real-world frequency of different leak counts (He & Garcia, 2009). After determining the number of leaks, the corresponding leaking nodes were selected randomly.

A leakage time step was then selected randomly from 1 to T, where T is the total number of time steps in the demand pattern. The selected leaks were active only during the sampled time step. This design exposed the models to leakage under different demand conditions while avoiding highly similar scenarios at consecutive time steps. Leak duration was not investigated in this study.

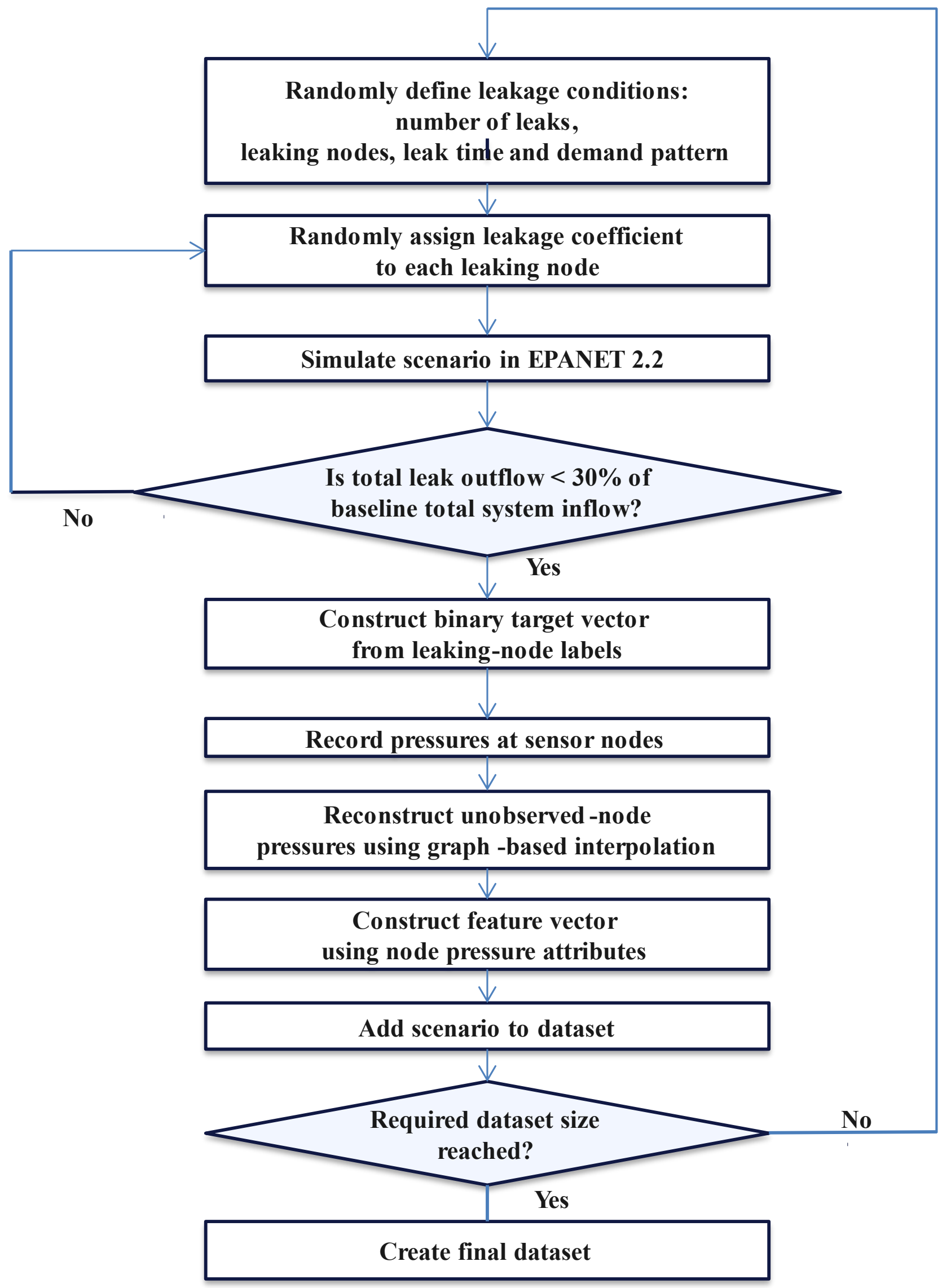

**Figure 1.** Workflow for generation of leakage scenarios and construction of the hydraulic dataset.

As shown in Figure 1, demand conditions were varied after selecting the leakage time step. To represent uncertainty and fluctuations in water consumption, the prescribed demand-pattern multiplier at the selected time step was multiplied by a random factor between 0.8 and 1.2. This range represents a maximum deviation of ±20% from the prescribed demand pattern, consistent with demand-uncertainty levels considered in previous WDN leakage studies (Tyagi et al., 2023; Li et al., 2025). The uniform distribution was selected to introduce bounded variability without systematically favoring either increases or decreases in demand.

Leakage magnitude was subsequently assigned independently to each leaking node using the EPANET 2.2 emitter formulation (Rossman et al., 2020):

$$Q = CP^{\gamma} \tag{1}$$

where $Q$, $C$, $P$, and $\gamma$ denote the leakage discharge, emitter coefficient, pressure at the emitter (leaking node), and emitter exponent, respectively. Assuming round-shaped leakage points, the emitter exponent was set to 0.5 (Mücke et al., 2023). To simulate different leak sizes, the emitter coefficient for each leakage was randomly assigned a value in the range $0 < C \leq 10$. Although the emitter coefficient can theoretically take any positive value, this range was selected based on engineering judgment to produce leaks of varying sizes.

After each hydraulic simulation, total leakage discharge was calculated as the sum of the emitter discharges from all leaking nodes. To avoid excessively severe leakage scenarios, a scenario was retained only when total leakage discharge was less than 30% of the baseline total system inflow at the corresponding time step, consistent with reported leakage levels in WDNs and previous leak-scenario generation studies (Ponti et al., 2024; Ajoodani et al., 2025). If this criterion was not satisfied, the leakage conditions were regenerated and the hydraulic simulation was repeated. This procedure continued until the required dataset size was reached.

### 2.2 Input and Target Construction

For each scenario, the target vector $y \in \{0,1\}^N$ contained one element for each leak-candidate node, where $N$ is the number of leak-candidate nodes. Each element $y_i$ indicated the leakage status of node $i$, with $y_i = 1$ for a leaking node and $y_i = 0$ otherwise. The input vector had the same dimension and contained one pressure-related feature for each leak-candidate node.

Only a small subset of the leak-candidate nodes was assumed to be equipped with pressure sensors, reflecting sparse monitoring conditions. Consequently, direct pressure measurements represented only part of the network-wide hydraulic state. Graph-based interpolation was therefore used to infer the hydraulic state at unobserved nodes from the available sensor measurements and network connectivity, providing a spatially complete input representation under sparse monitoring. Similar graph-based approaches have been used to reconstruct hydraulic-head or pressure fields from sparse measurements in WDNs (Hajgató et al., 2021; Irofti et al., 2022; Truong et al., 2024).

Reconstruction was performed using hydraulic head rather than pressure alone so that differences in node elevation were explicitly considered. For each sensor node $i$, the observed hydraulic head was calculated as

$$h_i = p_i + z_i \tag{2}$$

where $h_i$, $p_i$, and $z_i$ denote the hydraulic head, pressure head, and elevation of node $i$, respectively.

For the reconstruction step, network connectivity was represented by a weighted graph $G = (V, E)$, where $V$ denotes the set of nodes and $E$ denotes the set of connecting pipes. Each pipe was assigned a weight inversely proportional to its length:

$$w_{ij} = \begin{cases} \frac{1}{l_{ij}} & (i,j) \in E \\ 0 & \text{otherwise} \end{cases} \tag{3}$$

where $l_{ij}$ is the length of the pipe connecting nodes $i$ and $j$.

The weighted adjacency matrix $W$, weighted degree matrix $D$, and graph Laplacian $L$ were then defined as:

$$D_{ii} = \sum_{j} w_{ij}\,, \qquad L = D - W \tag{4}$$

The observed hydraulic heads at sensor nodes were imposed as fixed boundary values, while the heads at unobserved nodes were estimated by minimizing the weighted differences between connected nodes:

$$h = \min\left[\frac{1}{2}\sum_{(i,j)\in E} w_{ij}\left(h_i - h_j\right)^2\right], \quad \text{subject to } h_i = h_i^{\mathrm{obs}},\ i \in S \tag{5}$$

where $S$ denotes the set of sensor nodes.

After partitioning the graph Laplacian according to sensor and unobserved nodes:

$$L = \begin{bmatrix} L_{SS} & L_{SU} \\ L_{US} & L_{UU} \end{bmatrix} \tag{6}$$

where $U$ denotes the set of unobserved nodes, the hydraulic heads at the unobserved nodes were obtained as

$$\hat{\mathrm{h}}_U = -L_{UU}^{-1} L_{US} \mathrm{h}_S \tag{7}$$

The reconstructed pressure head at each unobserved node was subsequently calculated as

$$\hat{p}_i = \hat{h}_i - z_i, \qquad i \in U \tag{8}$$

After converting the reconstructed hydraulic heads to pressure, the resulting network-wide nodal pressure vector was used as the input feature vector for leak localization.

### 2.3 Graph Convolutional Network

A graph convolutional network (GCN), a class of graph neural networks (GNNs), was used for node-level leak localization (Kipf & Welling, 2017). The WDN was represented as a graph in which leak-candidate consumption nodes were represented as graph nodes and pipes as edges connecting hydraulically adjacent nodes. Each node was assigned the pressure-related feature constructed in Section 2.2, allowing the GCN to use both nodal hydraulic information and network connectivity during classification.

At each graph-convolution layer, the representation of a node is updated by aggregating information from the node itself and its neighboring nodes. Figure 2 illustrates this neighborhood-aggregation process. After one layer, a node incorporates information from its immediate neighbors; with additional layers, information can propagate from progressively more distant parts of the network. For example, in Figure 2, node A first receives information from directly connected nodes B, C, D, and E, while a second layer allows information originating from nodes G, H, I, and J to influence the representation of node A through the intermediate neighboring nodes.

This topology-aware information propagation is particularly useful under sparse monitoring because hydraulic information available at a limited number of locations can influence the learned representations of other nodes through the network structure. The final node representations are then used to estimate the leakage status of each leak-candidate node, allowing multiple nodes to be identified as leaking within the same scenario.

The number of graph-convolution layers (d) and the number of hidden neurons per layer (s) were treated as tunable hyperparameters. Increasing d allows information to propagate across progressively larger graph neighborhoods, whereas s controls the dimensionality of the learned node representations.

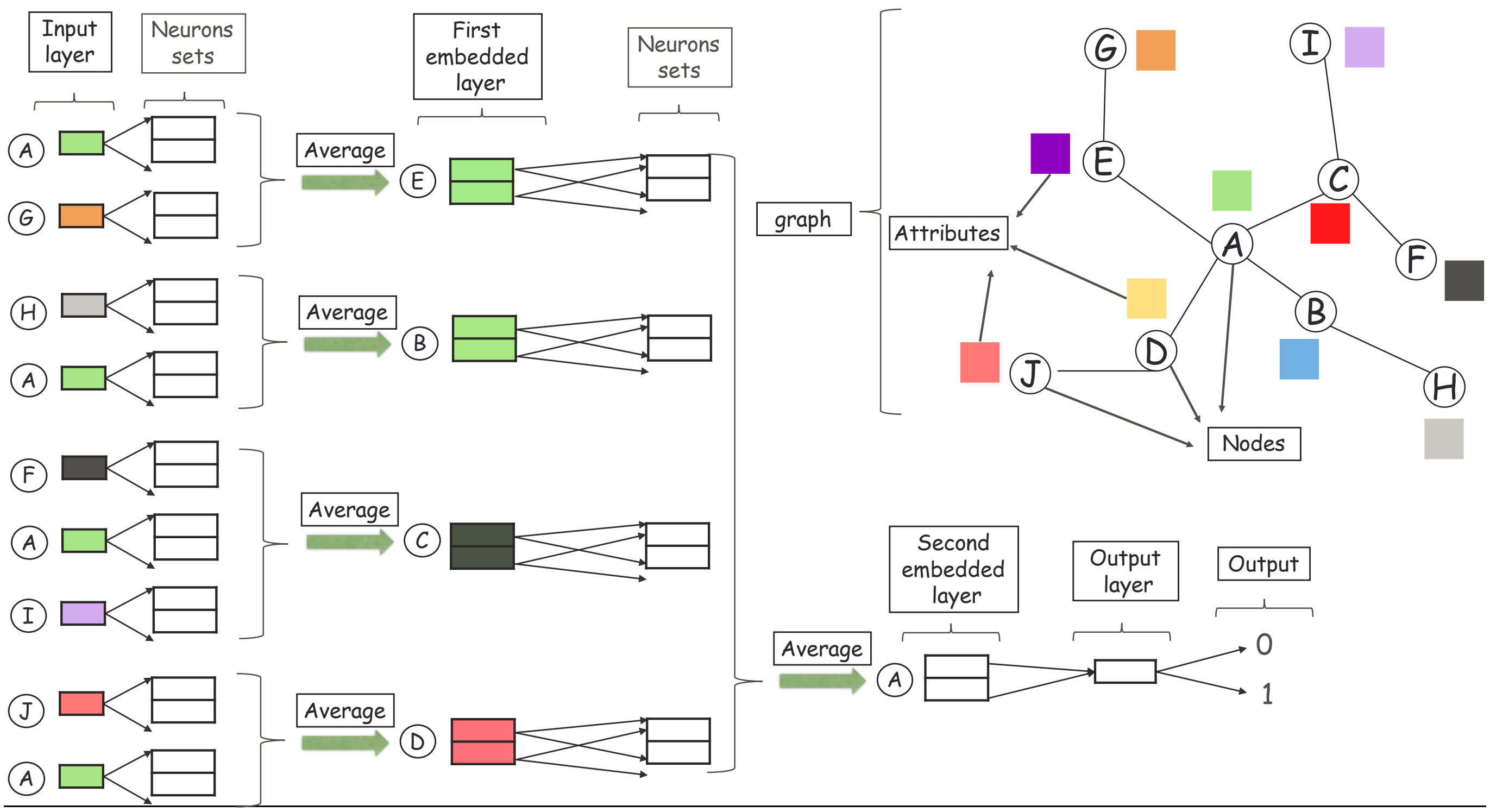

**Figure 2.** Neighborhood aggregation and information propagation in the GCN model.

### 2.4 Conventional Classifiers

MLP, SVM, KNN, and RF classifiers were developed alongside the GCN for performance comparison. All conventional classifiers used the pressure-related input features described in Section 2.2. All conventional classifiers were configured to produce one binary prediction for each leak-candidate node, consistent with the multilabel target representation used by the GCN.

The MLP is a feedforward neural network consisting of an input layer, one or more fully connected hidden layers, and an output layer. In this study, the number of hidden layers and the number of neurons per hidden layer were treated as tunable hyperparameters (Goodfellow et al., 2016).

The KNN classifier determines the class of a sample based on its K nearest training samples in the feature space. The number of neighbors, K, was treated as a tunable hyperparameter (Yiğit, 2013).

The SVM classifier identifies a decision boundary that maximizes the separation between classes. Its performance depends on the regularization parameter C and the selected kernel function. Accordingly, C, kernel type, and polynomial degree, when applicable, were treated as tunable hyperparameters (Patle & Chouhan, 2013).

The RF classifier combines predictions from multiple decision trees. The number of trees controls the ensemble size, whereas the maximum tree depth controls the complexity of individual trees. Both parameters were treated as tunable hyperparameters (Biau & Scornet, 2016).

**2.5 Model Training and Evaluation**

The GCN model was trained using the binary cross-entropy (BCE) loss function together with the Adam optimization algorithm (Kingma & Ba, 2015). For each model configuration, training continued until the predefined training-convergence criterion was satisfied. Training convergence was defined as an improvement in the training loss of less than 1% over 10 consecutive training iterations. The BCE loss was calculated over all leak-candidate nodes and training scenarios as follows:

$$L_{BCE} = -\frac{1}{mn}\sum_{i=1}^{n}\sum_{j=1}^{m}\left[\, Y_{i_j}.\log\left(\hat{Y}_{i_j}\right) + \left(1 - Y_{i_j}\right).\log\left(1 - \hat{Y}_{i_j}\right)\right] \tag{9}$$

Where $n, m\ Y_{i_j}$, $\hat{Y}_{i_j}$, and $L_{BCE}$ denote the number of nodes, the number of scenarios, the actual value of the $i_{th}$ node of the $j_{th}$ scenario, the predicted value of the $i_{th}$ node of the $j_{th}$ scenario and the loss function value, respectively.

For each investigated dataset size, the generated scenarios were randomly divided into development and test subsets using an 80/20 split. The 80% development subset was used for model training and hyperparameter selection, while the remaining 20% was retained as a held-out test subset. Hyperparameters were selected using five-fold cross-validation within the development subset (Arlot & Celisse, 2010). In each cross-validation iteration, four folds were used for training and the remaining fold for validation, such that each fold served once as the validation subset. The same data split and cross-validation procedure were applied to the GCN and all conventional classifiers to ensure a consistent comparison.

The mean F1-score across the five validation folds was used to select the optimal hyperparameter configuration. The selected configuration was then retrained using the complete 80% development subset and evaluated on the held-out 20% test subset. The hyperparameters considered for the GCN and conventional classifiers are presented in Table 1.

To investigate the amount of training data required for stable model performance, the complete cross-validation procedure was repeated using progressively larger datasets. Dataset enlargement was continued while the mean cross-validation F1-score showed meaningful improvement. A learning-curve plateau was considered to have been reached when the absolute increase in mean cross-validation F1-score was less than 0.01 over two consecutive dataset-size increments. This plateau was used as the stopping criterion for further scenario generation. Importantly, the held-out test subsets were not used to determine this stopping point.

Precision, recall, F1-score, and accuracy were used to evaluate node-level leak-localization performance, as defined in Equations (10)– (13) (Yacouby & Axman, 2020).

$$\text{Precision} = \frac{\text{TP}}{\text{TP} + \text{FP}} \tag{10}$$

$$\text{Recall} = \frac{\text{TP}}{\text{TP} + \text{FN}} \tag{11}$$

$$\text{F1} - \text{score} = \frac{2 \times \text{Precision} \times \text{Recall}}{\text{Precision} + \text{Recall}} \tag{12}$$

$$\text{Accuracy} = \frac{\text{TP} + \text{TN}}{\text{TP} + \text{TN} + \text{FP} + \text{FN}} \tag{13}$$

Here, TP denotes the number of leaking nodes correctly classified as leaking, FP the number of non-leaking nodes incorrectly classified as leaking, TN the number of non-leaking nodes correctly classified as non-leaking, and FN the number of leaking nodes incorrectly classified as non-leaking. The reported metrics were calculated for each leak-candidate node and then averaged across all nodes.

**Table 1.** Candidate hyperparameter values for each model.

| **Model** | **Hyperparameters** | **Candidates** |
|---|---|---|
| GCN | d (number of graph-convolution layers) | {1, 2, 3, 4, 5, 6, 7} |
| | s (hidden neurons per graph-convolution layer) | {8, 16, 32, 64, 128, 256} |
| MLP | Number of hidden layers | {1, 2, 3, 4, 5, 6} |
| | Number of neurons per hidden layer | {16, 32, 64, 128, 256, 512} |
| KNN | K | {1, 2, 3, ..., 25} |
| SVM | C | {0.1, 1, 100} |
| | Kernel | {linear, poly, rbf} |
| | Degree (poly kernel only) | {2, 3} |

| Model | Hyperparameters | Candidates |
|---|---|---|
| RF | Number of trees | {100, 200, 300, …, 1000} |
| | Maximum tree depth | {10, 20, 30, 40, 60, None} |

### 2.6 Benchmark Networks and Experimental Setup

The proposed framework was evaluated using two benchmark WDNs of different sizes. The Hanoi network, originally introduced by Fujiwara and Khang (1990), consists of 31 nodes, 34 pipes, and one reservoir (Figure 3); its hydraulic and demand data were adopted from the benchmark configuration used by Lučin et al. (2021). The larger NET3 network is distributed with EPANET and contains 92 nodes, 117 pipes, two reservoirs, and three tanks (Rossman et al., 2020) (Figure 4). Using networks of substantially different sizes provides a basis for evaluating the applicability of the proposed approach to increasingly complex WDNs.

To represent sparse monitoring conditions, pressure measurements were assumed to be available at four nodes in the Hanoi network and eleven nodes in the NET3 network. Sensor locations were selected to provide spatial and hydraulic coverage of different parts of each network, including locations near water sources, terminal branches, elevated regions, and hydraulically remote areas.

For the Hanoi network, pressure sensors were assigned to nodes 2, 13, 22, and 31. Node 2 is located near the reservoir, nodes 13 and 22 are located at terminal branches, and node 31 represents a hydraulically remote part of the network.

For the NET3 network, pressure sensors were assigned to nodes 10, 15, 20, 40, 50, 60, 131, 153, 219, 243, and 253. Nodes 10 and 60 are located near the lake and river sources, respectively; nodes 15, 131, 219, 243, and 253 are located at terminal branches; and nodes 10, 20, 40, 50, and 153 represent high-elevation or local-elevation maxima. Nodes 20, 40, and 50 are also located near tanks 3, 1, and 2, respectively, while node 219 has the greatest shortest-path distance to its nearest water source and therefore represents a hydraulically remote region of the network.

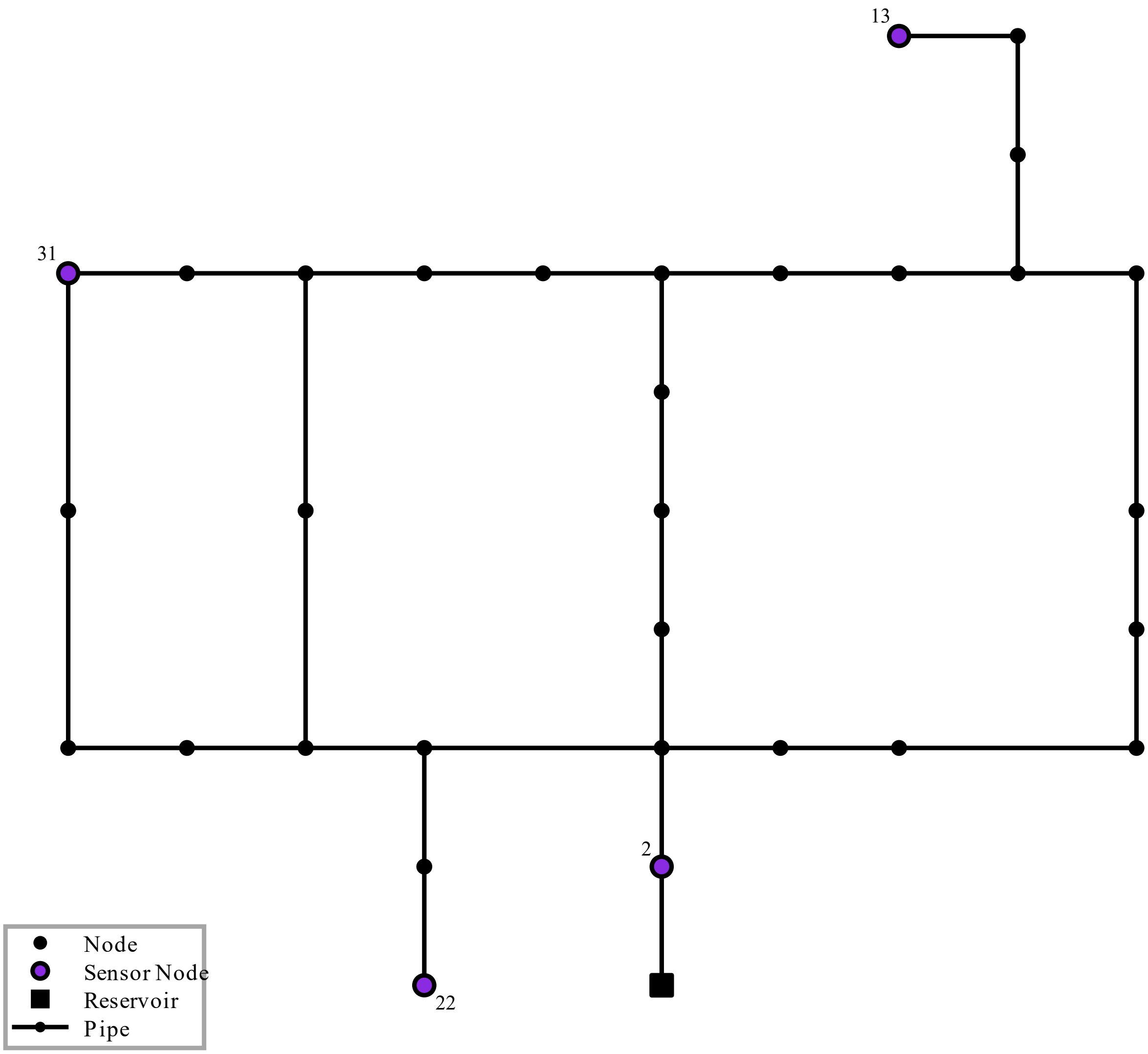

**Figure 3.** Hanoi WDN with 31 demand nodes, 34 pipes, one reservoir, and four pressure-sensor nodes.

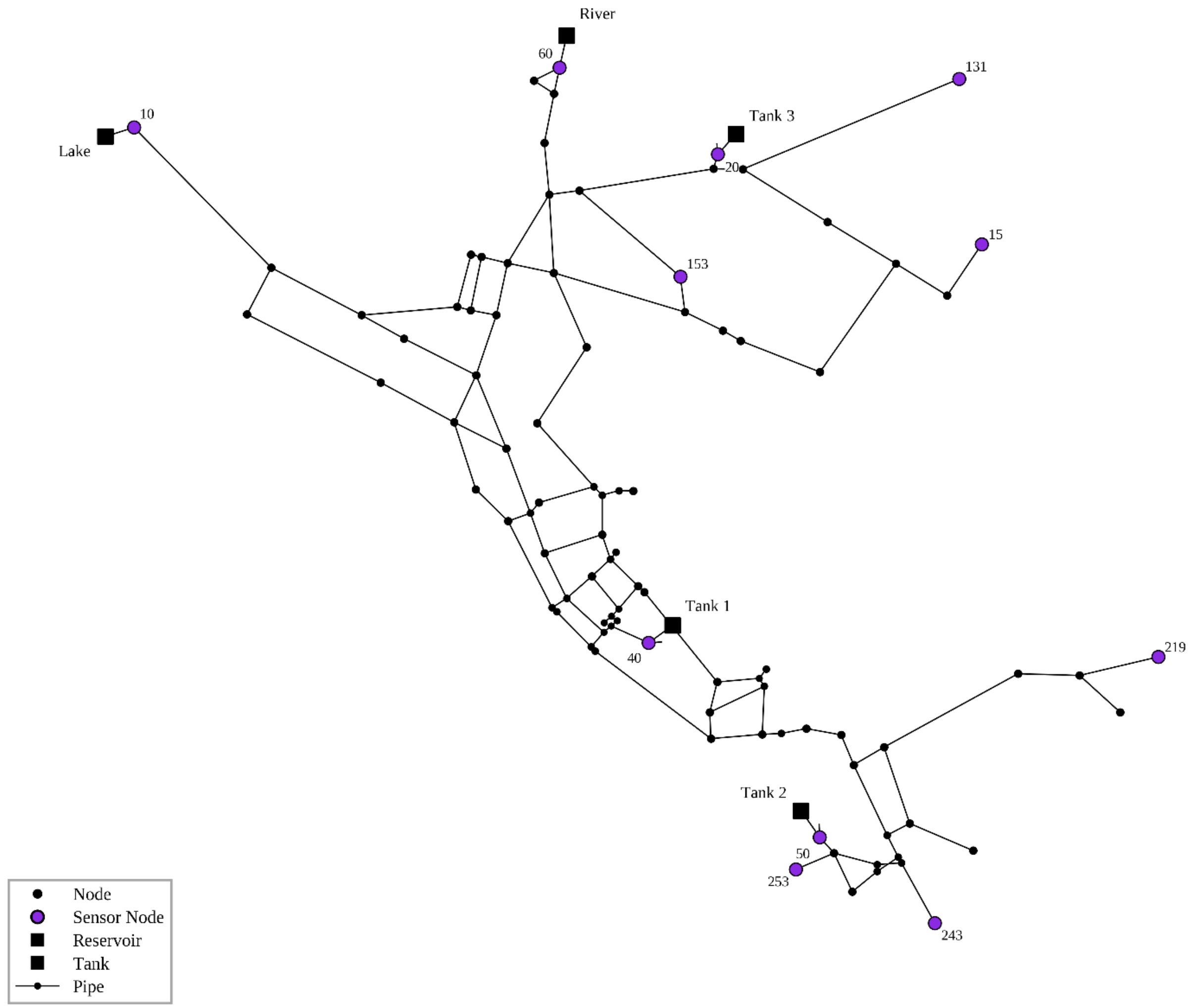

**Figure 4.** NET3 WDN with 92 consumption nodes, 117 pipes, two reservoirs, three tanks, and eleven pressure-sensor nodes.

## 3. Results and Discussion

The results are presented in two parts. First, the overall performance of the GCN and the conventional classifiers is evaluated as the number of generated scenarios increases. Second, node-level classification accuracy is examined in relation to the average shortest-path distance to pressure sensors.

### 3.1 Model Performance and Effect of Dataset Size

The GCN, RF, MLP, SVM, and KNN classifiers were evaluated on the Hanoi and NET3 networks using different numbers of generated scenarios, ranging from 2000 to 22000 for Hanoi and from 4000 to 44000 for NET3 (Figures 5 and 6). Each scenario contained a randomly selected number of simultaneous leaks, ranging from one to the total number of leak-candidate nodes, and the leaking nodes were selected randomly.

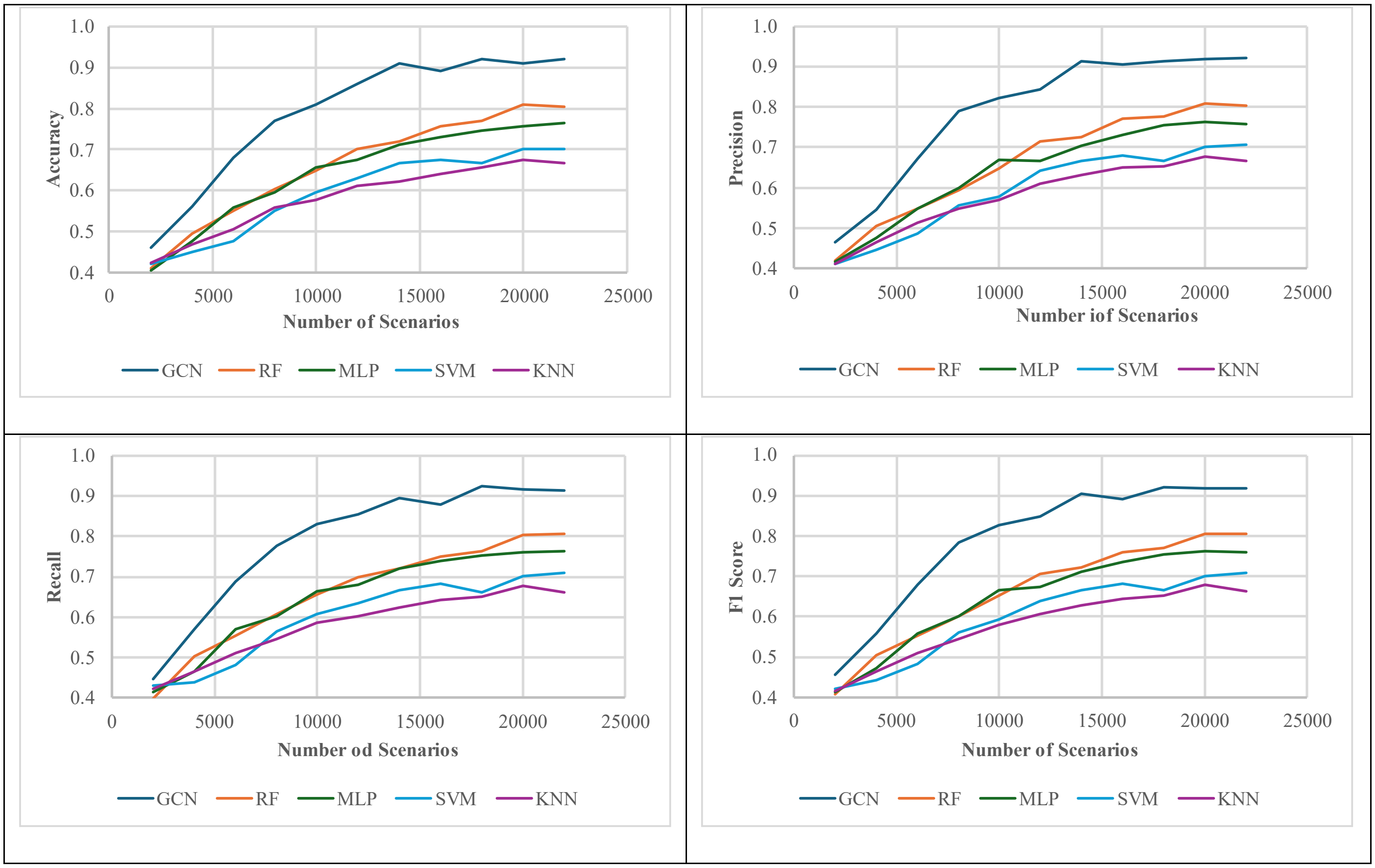

**Figure 5.** Performance of the GCN, RF, MLP, SVM, and KNN models as the number of scenarios increases for the Hanoi network.

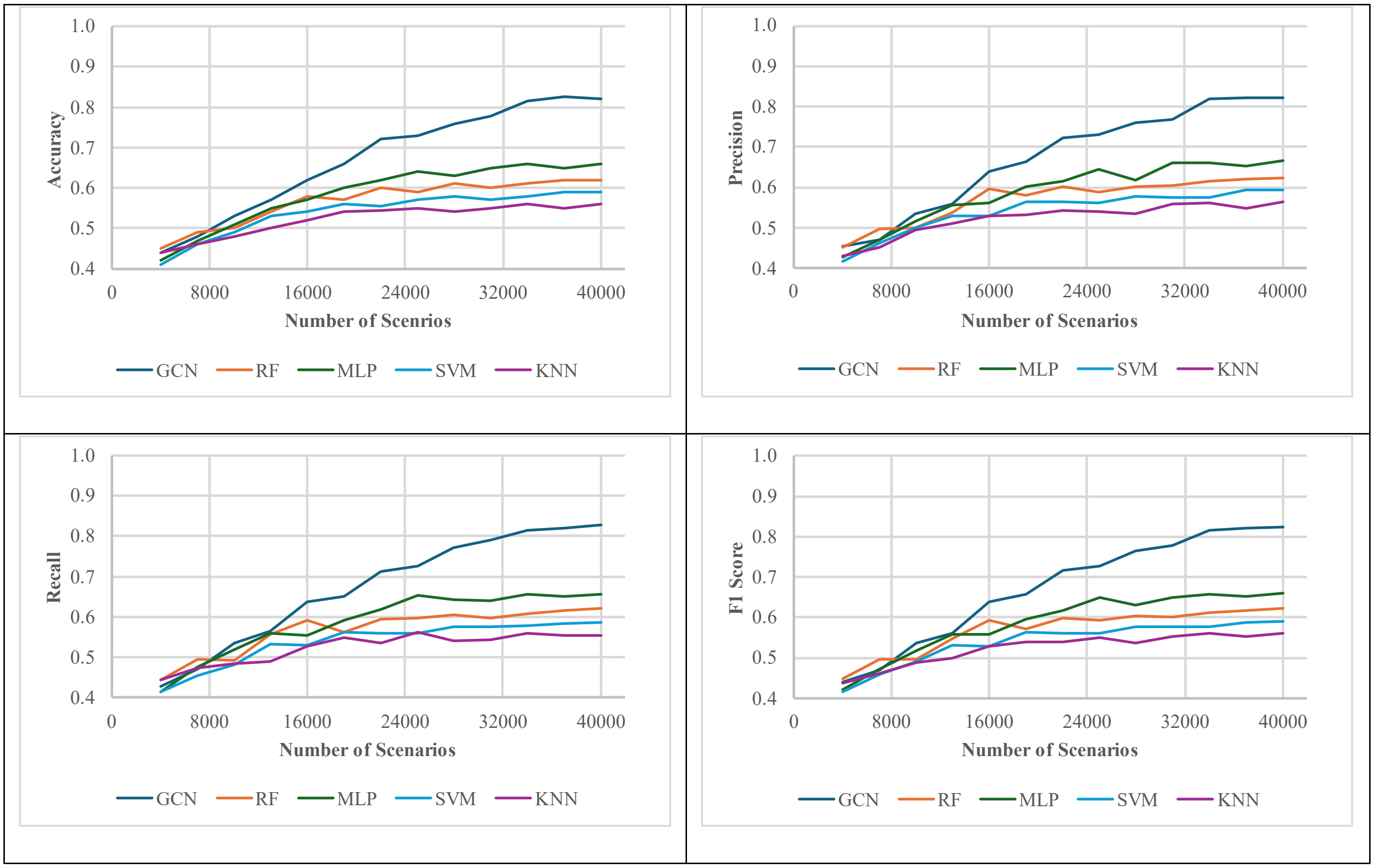

**Figure 6.** Performance of the GCN, RF, MLP, SVM, and KNN models as the number of scenarios increases for the NET3 network.

The GCN outperformed the conventional classifiers across the evaluated metrics. All models used the same reconstructed pressure-related input features; however, the GCN additionally incorporated network topology during classification through graph-based neighborhood aggregation. This additional topological information explains its stronger performance relative to the conventional classifiers.

The scale of the sampled scenario space can be evaluated relative to the total number of possible leak-location combinations. For the Hanoi and NET3 networks,

$$C_{\text{Hanoi}} = 2^{31} - 1 = 2{,}147{,}483{,}647 \quad (14)$$

$$C_{\text{NET3}} = 2^{92} - 1 \approx 4.95 \times 10^{27} \quad (15)$$

where C denotes the total number of possible leak-location combinations Accordingly, 22,000 Hanoi scenarios represent

$$\frac{22,000}{2^{31}-1} \times 100 \approx 0.0010\%,$$

of all the possible leak-location combinations, while approximately 42,000 NET3 scenarios represent only

$$\frac{42,000}{2^{92}-1} \times 100 \approx 8.5 \times 10^{-22}\%.$$

These results show that effective localization can be achieved using only a very small sampled fraction of the combinatorial leak-location space, supporting the scalability of the proposed formulation across the tested network sizes.

Compared with Hanoi, all models showed lower performance on the larger and more complex NET3 network. However, the GCN's advantage over the conventional classifiers became more pronounced, suggesting that topology-aware learning provides greater benefits as WDN size and complexity increase.

### 3.2 Node-Level Accuracy and Sensor Proximity

For the Hanoi network, the GCN achieved an overall accuracy of 0.92 on the held-out test subset of the 22,000-scenario dataset. . However, node-level accuracy ranged from 0.84 to 0.97 The relationship between each node's average shortest-path distance (ASPD) from the pressure-sensor nodes and its classification accuracy was therefore examined (Figure 7; Table 2). As ASPD decreased, node-level classification accuracy increased.

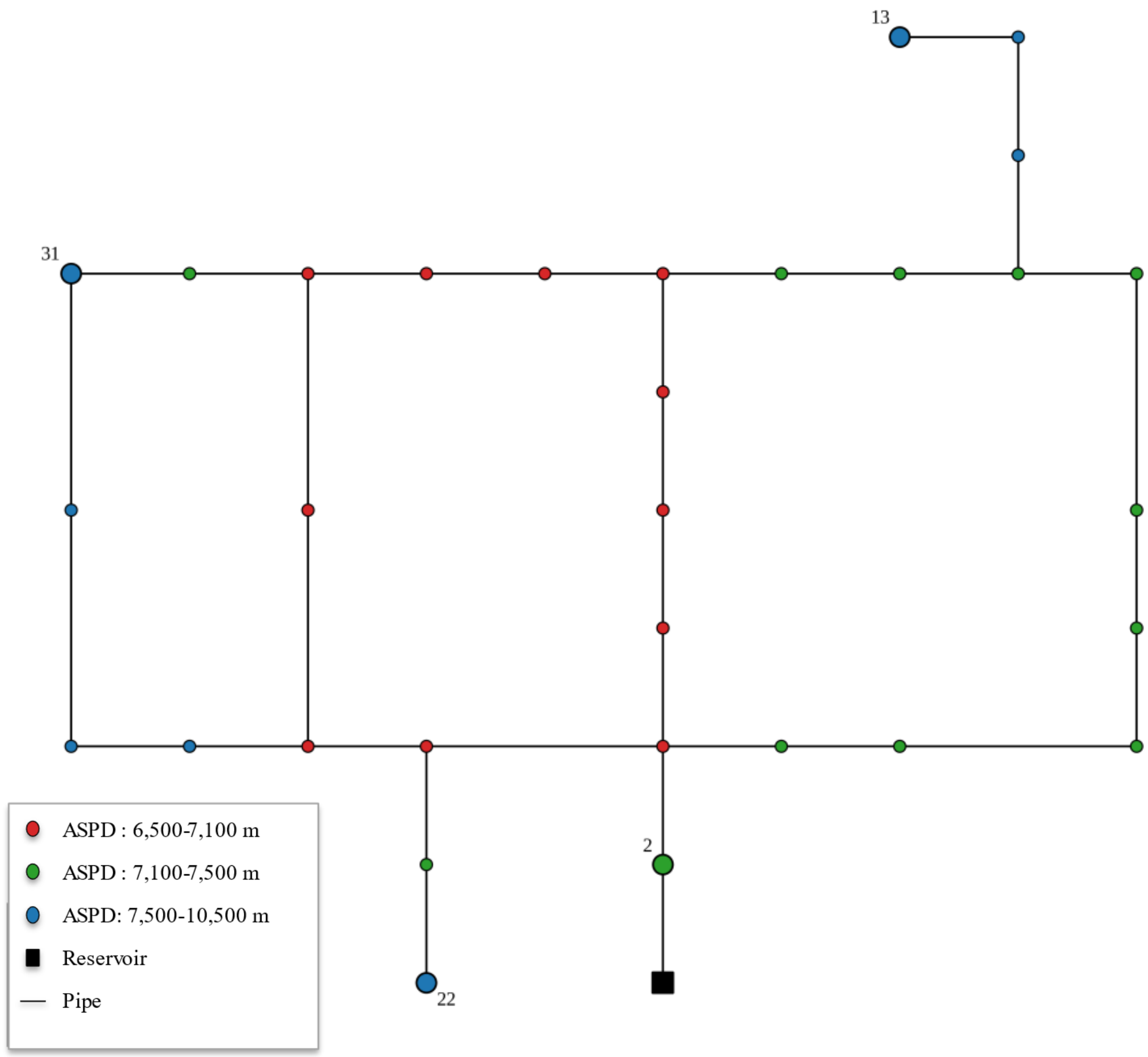

**Figure 7.** Classification of Hanoi network nodes according to their ASPD.

**Table 2.** Relationship between ASPD and accuracy for the Hanoi network.

| **ASPD from sensor nodes** | **Average accuracy** |
|---|---|
| (6500, 7100) | 0.94 |
| (7100, 7500) | 0.91 |
| (7500, 10500) | 0.86 |

The same analysis was applied to the NET3 network, where a similar pattern was observed. The reported overall accuracy for 40000 scenarios was 0.82 As ASPD decreased, node-level classification accuracy increased, as shown in Figure 8 and Table 3.

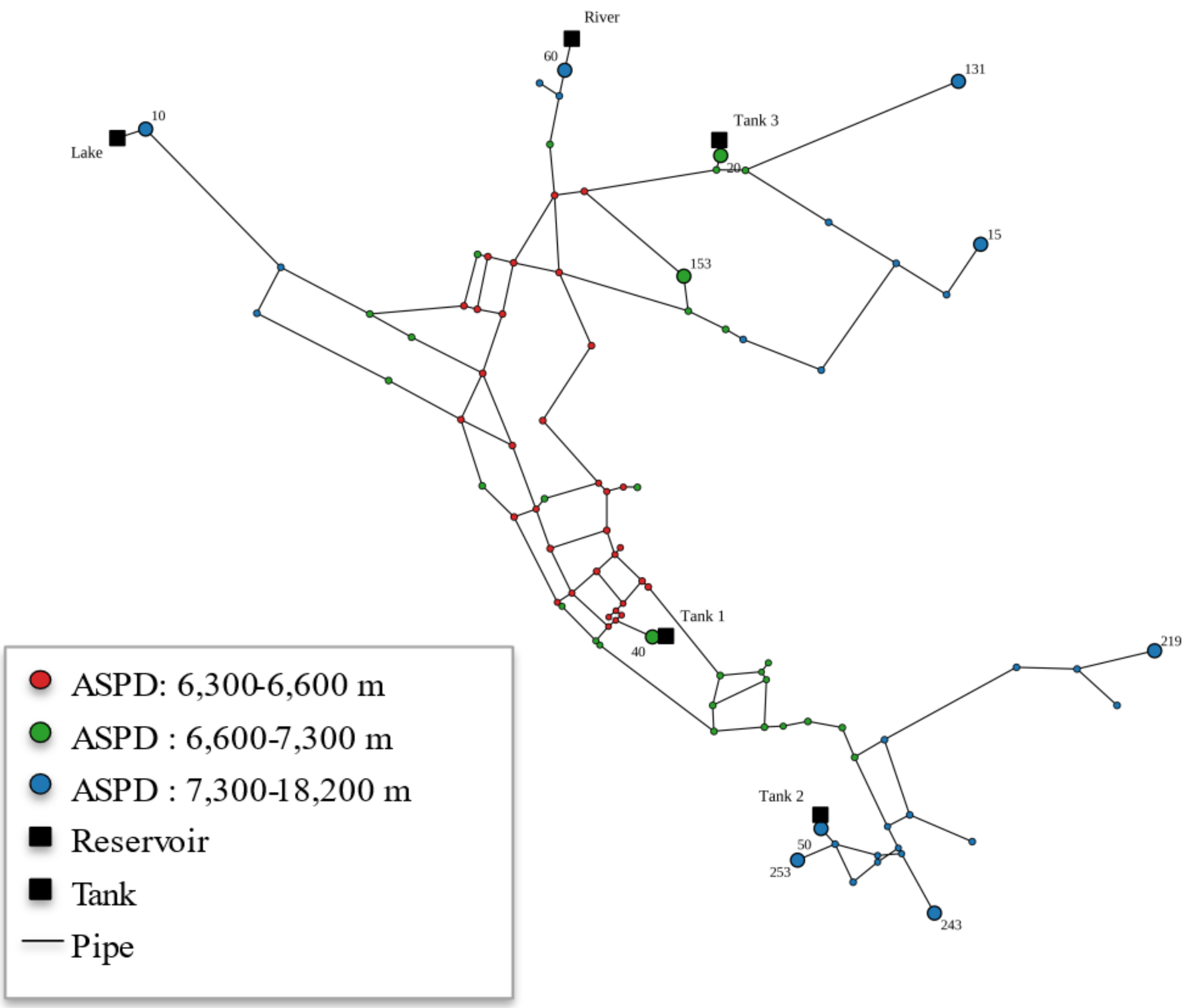

**Figure 8.** Classification of NET3 network nodes according to their ASPD.

**Table 3.** Relationship between ASPD and accuracy for the NET3 network.

| ASPD from sensor nodes | Average accuracy |
|---|---|
| (6300, 6600) | 0.87 |
| (6600, 7300) | 0.80 |
| (7300, 18200) | 0.75 |

The ASPD intervals in Tables 2 and 3 were selected to obtain approximately balanced numbers of nodes across the distance ranges, reducing the influence of highly unequal group sizes on the grouped accuracy analysis.

## Conclusion

This study developed a GCN-based framework for localizing a variable number of simultaneous leaks in WDNs under sparse monitoring. The GCN was evaluated alongside MLP, SVM, RF, and KNN classifiers and achieved consistently stronger leak-localization performance. By incorporating network topology through graph convolution, the proposed model can use spatial relationships between nodes in addition to the hydraulic information.

The proposed framework has three main features. First, it does not require the number of leaks to be specified in advance. Second, it can be trained using a sampled subset of leakage scenarios rather than exhaustively generating all possible leak-location combinations Third, it operates under sparse pressure-sensor coverage by reconstructing pressure information at unmonitored nodes using network connectivity.

The framework was evaluated under several conditions relevant to operational WDNs, including variable demand, random leak locations, variable leak magnitudes, a variable number of simultaneous leaks, and sparse pressure-sensor coverage. Considering these conditions provides a more practical assessment of model performance than evaluations based on fixed hydraulic or leakage conditions. The results for both Hanoi and NET3 therefore support the practical applicability of the proposed framework under the conditions considered in this study. Future work should investigate additional hydraulic information, such as flow measurements, and more sample-efficient training strategies to further improve leak-localization performance and reduce data requirements.

**Data Availability Statement:** Data and materials supporting the results of this paper are available upon reasonable request.

**Funding:** This research received no external funding.

**Conflicts of Interest:** No potential conflicts of interest were reported by the authors.